\documentclass{iac}

\begin{document}

\IACconference{75}
\IAClocation{Milan, Italy}
\IACdates{14-18 October}
\IACyear{2024}
\IACpapernumber{A3}

\title{Modular pipeline for small bodies gravity field modeling: an efficient representation of variable density spherical harmonics coefficients}

\IACauthor*{Antonio Rizza}{PhD Candidate,  Department of Aerospace Science and Technology (DAER), Politencico di Milano, \mbox{Via La Masa 34}}{antonio.rizza@polimi.it}
\IACauthor{Carmine Buonagura}{PhD Candidate, Department of Aerospace Science and Technology (DAER), Politencico di Milano, \mbox{Via La Masa 34}}{carmine.buonagura@polimi.it}
\IACauthor{Paolo Panicucci}{Assistant Professor, Department of Aerospace Science and Technology (DAER), Politencico di Milano,\mbox{Via La Masa 34}}{paolo.panicucci@polimi.it}
\IACauthor{Francesco Topputo}{Full Professor, Department of Aerospace Science and Technology (DAER), Politencico di Milano, \mbox{Via La Masa 34}}{francesco.topputo@polimi.it}

\abstract{Proximity operations to small bodies, such as asteroids and comets, demand high levels of autonomy to achieve cost-effective, safe, and reliable Guidance, Navigation and Control (GNC) solutions. Enabling autonomous GNC capabilities in the vicinity of these targets is thus vital for future space applications. However, the highly non-linear and uncertain environment characterizing their vicinity poses unique challenges that need to be assessed to grant robustness against unknown shapes and gravity fields. In this paper, a pipeline designed to generate variable density gravity field models is proposed, allowing the generation of a coherent set of scenarios that can be used for design, validation, and testing of GNC algorithms. The proposed approach consists in processing a polyhedral shape model of the body with a given density distribution to compute the coefficients of the spherical harmonics expansion associated with the gravity field. To validate the approach, several comparison are conducted against analytical solutions, literature results, and higher fidelity models, across a diverse set of targets with varying morphological and physical properties. Simulation results demonstrate the effectiveness of the methodology, showing good performances in terms of modeling accuracy and computational efficiency. This research presents a faster and more robust framework for generating environmental models to be used in simulation and hardware-in-the-loop testing of onboard GNC algorithms.}

\maketitle

\section{Introduction}
In the last twenty years, the space sector has experienced an unprecedented growth, marked by significant advancements and achievements both concerning Earth's orbiting and deep-space missions. The recent growing interest in small solar system bodies such as asteroids and comets for scientific inspection, exploitation of resources, and planetary defense reasons is pushing the development of innovative engineering solutions to better investigate these celestial bodies. Ground-based observations allow preliminary characterizations of small bodies in terms of bulk properties, such as mass and shape, orbit and rotational state. The major limitation of this methodology is the signal to noise ratio \cite{buonagura2024} which is acceptable only when the target is relatively close to the Earth  \cite{scheeres2016orbital}. A drastic improvement in the body characterization can be obtained with in-situ observations with the use of specialized and instrumented probes. Several missions successfully performed proximity operations to these bodies such as the Near Earth Asteroid Rendezvous (NEAR) Shoemaker \cite{NEARMissionDesign}, Dawn \cite{Dawn_RAYMAN2020233}, the Origins, Spectral Interpretation, Resource Identification, Security, Regolith Explorer (OSIRIS-REx) \cite{osiris-rex}, Hayabusa \cite{hayabusa}, \mbox{Hayabusa2 \cite{hayabusa2}}, \mbox{Rosetta \cite{RosettaOverview}}, and the Double Asteroid Redirection Test (DART)\cite{dart,cheng2023momentum}. Nowadays, space exploration is witnessing a transition towards the use of CubeSats, miniaturized platforms standardized in size and form factor, for the systematic exploration of the Solar System \cite{walker2018deep}. The idea is to use these platforms performing riskier tasks and operating in a cooperative multi-agent framework while cooping with limited resources \cite{CubeSatsAutonomy}. \\

The dynamics characterizing small bodies proximity operations is highly chaotic with solar radiation pressure and non-spherical gravity effects compromising the very existence of stable closed orbits. An accurate modeling of this environment becomes crucial for a safe trajectory design and to guarantee the satisfaction of mission objectives. Moreover, on-board operations requires fast computation to coop with limited available resources. A good compromise between accuracy and efficiency in modeling the gravity field, is typically found in an high-order expansion of the gravitational potential in the form of spherical harmonics. This model is typically valid only outside the Brillouin sphere of the target \cite{scheeres2016orbital}, because of the convergence properties of the Legendre polynomials, used as functional base for the expansion \cite{Takahashi}. An alternative formulation resolving the convergence issue inside the Brillouin sphere is the polyhedral gravity model proposed in Werner and Scheeres \cite{werner1996exterior}. This model computes the gravity field from a constant density polyhedron in an analytical form. Given that the polyhedron is usually computed from images, errors are present in the polyhedral model. The sensitivity of the gravity field to perturbation in the polyhedral  shape are investigated by Bercovici et al. \cite{bercovici2020analytical}, to assess the coupling between shape and gravity errors for Werner and Scheeres's \cite{werner1996exterior} model. One of the main drawback of the polyhedral gravity model is that it is associated with a constant density distribution inside the body which sometimes results in contrast with gravity measurements \cite{ScheeresHeterogeneousBennu,FERRARI2022114914}. Even a small density variation may induce large trajectory deviation with respect to a nominal path, leading the spacecraft to miss its target goals and potentially enter on impact or escape trajectories. A quantitative example of this deviation is discussed later in this work using the variable density gravity model generated with the proposed approach.   \\

Motivated by the need of comparing spherical harmonics coefficients from orbit determination and the ones gathered from constant-density shapes, Werner \cite{WERNER19971071} shows how the spherical-harmonics coefficients can be analytically retrieved from a general shape polyhedral with uniform density model of the asteroid combining recursive formula with trinomial algebra.  The sensitivity of the spherical harmonics coefficients to shape variations is investigated by Panicucci et al. \cite{panicucci2020uncertainties} to understand whether the shape can be a major source of error when comparing orbit determination coefficients to shape-deduced coefficients (e.g., see \cite{ScheeresHeterogeneousBennu})  To improve the fidelity of the forward gravity modeling of non-constant density polyhedron, Chen et al. \cite{chen2019spherical} extends Werner \cite{WERNER19971071}  to variable density under the assumption of trinomial density distribution. However, this hypothesis force the density field to be continuous inside the body forbidding the existence of density jumps. This may not be the case when the asteroid is formed by two different parent bodies or when denser nuclei and internal geological structures are considered.\\

Another widely-used gravity field is mascon model, introduced to model gravity anomalies on the Moon \cite{arkani1998lunar}. Mascon models \cite{VariableDensityMascon,tardivel2016limits} are typically used to model non-uniform mass distribution. An interesting comparison among different gravity models, spherical harmonics, mascon and polyhedral, is provided by Werner and Scheeres \cite{werner1996exterior} for asteroid (4769) Castalia. To the authors knowledge there is currently no formulation of the spherical harmonics expansion under arbitrary variable density distribution and shape of the asteroid. This paper proposes a pipeline for retrieving such gravity model by performing analytical integration over a radially discretized polyhedron. This approach falls thus in between the methodologies proposed by Werner and Cheng \cite{WERNER19971071,chen2019spherical} and the classical Mascon approach. \\

The paper is structured as follows: Section \ref{sec: Gravity field modeling} describes the general formulation of the gravity field by means of a spherical-harmonics expansion, Section \ref{sec: Normalized coefficients computation} presents the methodology to compute the coefficients of the expansion with non-uniform density, Section \ref{sec: Validation} applies the methodology to benchmark cases to test its performances and, finally, Section \ref{sec: Conclusions} summarizes the major findings. 

\section{Gravity field modeling}\label{sec: Gravity field modeling}
The acceleration due to the central gravity field, expressed in the asteroid fixed frame $\mathcal{B}$, can be computed as the gradient of the gravitational potential $U\left(\vec{x}\right)$

\begin{equation}
    \vec{a}_{\mathcal{B}} = \nabla U
    \label{eq: central gravity acceleration}
\end{equation}

Being due to a conservative field, the potential has to satisfy the Laplace equation outside the body, i.e.

\begin{equation}
    \nabla^2 U \left(\vec{x}\right) = 0
    \label{eq: Laplace equation}
\end{equation}

A known solution to the Laplace equation, valid outside the Brouillon \mbox{sphere \cite{scheeres2016orbital}} of the asteroid, is expressed in spherical coordinates through an infinite expansion of the potential $U\left(r,\lambda,\phi\right)$ into a series of spherical harmonics projected onto a function space spanned by the associated Legendre \mbox{polynomials \cite{vallado2001fundamentals}}. The coordinates $r$, $\lambda$ and $\phi$ are respectively the radial distance from the asteroid, the longitude and the latitude of the spacecraft. To avoid numerical errors and improve accuracy for higher order terms this expansion is typically used in the normalized form

\begin{equation}
\begin{aligned}
    	\hspace{-3.8 mm} U  =\dfrac{\mu}{r} \sum_{n=0}^{\infty} \sum_{m=0}^{n} \left(\dfrac{R_{0}}{r}\right)^n \hspace{-1 mm} \Pbar{n}{m}{u}\left(\Cbar{n}{m} \cos\left(m \lambda\right)+\right.\\ +\left.\Sbar{n}{m} \sin \left(m \lambda\right) \right)
     \end{aligned}
	\label{eq: harmonic expansion normalized}
\end{equation}

where $u = \sin\left(\phi\right)$. The function $\Pbar{n}{m}{u}$ refers to the normalized associated Legendre polynomials, while, the scaling factor $R_0$ is a reference distance used to compute the normalized coefficients $\Cbar{n}{m}$, and $\Sbar{n}{m}$. The polynomials $\Pbar{n}{m}{u}$ are expressed as a function of the Legendre polynomials $P_n$ as \cite{vallado2001fundamentals} \\
\begin{equation}
	\Pbar{n}{m}{u} = N_{nm}\p{1-u^2}^{\dfrac{m}{2}} \dfrac{d^m P_n}{du^m}
	\label{eq: fully-normalized associated legendre polynomial}
\end{equation}
with 
\begin{equation}
	N_{nm} = \sqrt{\dfrac{\left(n-m\right) !  \left(2 n +1\right) \left(2-\delta\left(m\right) \right)}{\left( n+m\right) !}}
	\label{eq: normalization coefficients}
\end{equation}

and $\delta\left(\cdot\right)$ indicating the Dirac delta operator. 

\begin{equation}
	\delta \left(m\right)  = \left\{ 
	\begin{array}{ c l }
		1 &  m = 0 \\
    0 & \textrm{ otherwise}
	\end{array}
	\right.
	\label{eq: Dirac delta}
\end{equation}

To speed up the computation of the fully-normalized Legendre polynomials in Equation \ref{eq: fully-normalized associated legendre polynomial} the recursive formula introduced by Rapp \cite{rapp1982fortran} are used. The sequence is anchored on the first three elements as

\begin{equation}
	\begin{aligned}
		& \Pbar{0}{0}{u} = 1  &&  \\ \\
		& \Pbar{1}{0}{u} = \sqrt{3} u &&\\ \\
		& \Pbar{2}{2}{u} = \sqrt{3} \sqrt{1-u^2} && \\
	\end{aligned}
    \label{eq: recursive formula for the computation of P_nm anchor}
\end{equation}

and then the terms are recursively computed as in Equation \ref{eq: recursive formula for the computation of P_nm}.

\begin{equation}
	\begin{aligned}
        \textrm{For } m  = &n,  \textrm{  } n > 2,\\ \\
		\Pbar{n}{n}{u} = & \sqrt{\dfrac{2n+1}{2n}} \sqrt{1-u^2} \Pbar{n-1}{n-1}{u}\\ \\
        \textrm{for } m = &n-1,\\ \\
\displaystyle		\Pbar{n}{n-1}{u} =& \sqrt{2n+3} u \Pbar{n-1}{n-1}{u}\\ \\
        \textrm{for } m < &n-1,\\ \\
		\Pbar{n}{m}{u} =& \Gamma_{n,m} \Pbar{n-1}{m}{u} +\\&- \dfrac{\Gamma_{n,m}}{\Gamma_{n-1,m}} \Pbar{n-2}{m}{u} 
	\end{aligned}
	\label{eq: recursive formula for the computation of P_nm}
\end{equation}

with

\begin{equation}
	\Gamma_{n,m} = \sqrt{\dfrac{\left(2n+1\right) \left(2n-1\right)}{\left(n-m\right) \left(n+m\right)}}
	\label{eq: gamma_nm}
\end{equation}

Expressing the spacecraft position in Cartesian coordinates, i.e. $\vec{r}_{\mathcal{B}} = x \vec{i}+y \vec{j}+z \vec{k}$, the gradient of the pontetial can be expressed as \cite{vallado2001fundamentals}

\begin{equation}
	\begin{aligned}
		&\vec{a}_{\mathcal{B}} = \nabla U =\left[\dfrac{\partial U}{\partial \vec{r}}\right]^T = \\ \\
        = & \dfrac{\partial U}{\partial r} \left[\dfrac{\partial r}{\partial \vec{r}} \right]^T +  \dfrac{\partial U}{\partial \lambda} \left[\dfrac{\partial \lambda}{\partial \vec{r}} \right]^T +  \dfrac{\partial U}{\partial \phi} \left[\dfrac{\partial \phi}{\partial \vec{r}} \right]^T = \\ \\
   = & \left[ \left(\dfrac{1}{r} \dfrac{\partial U}{\partial r} - \dfrac{z}{r^2 \eta} \dfrac{\partial U }{\partial \phi}\right) x - \left(\dfrac{1}{\eta^2} \dfrac{\partial U}{\partial \lambda}\right) y\right] \vec{i} + \\ \\ 
		+& \left[ \left(\dfrac{1}{r} \dfrac{\partial U}{\partial r} - \dfrac{z}{r^2 \eta} \dfrac{\partial U }{\partial \phi}\right) y + \left(\dfrac{1}{\eta^2} \dfrac{\partial U}{\partial \lambda}\right) x\right] \vec{j} + \\ \\
		+& \left[ \dfrac{1}{r} \dfrac{\partial U}{\partial r} z + \dfrac{\eta}{r^2} \dfrac{\partial U}{\partial \phi}\right] \vec{k}
	\end{aligned}
	\label{eq: gradient of the potential 2}
\end{equation}

with $\eta = \sqrt{x^2+y^2}$. The partial derivatives $\dfrac{\partial U}{\partial r}$ and $\dfrac{\partial U}{\partial \lambda}$ are given by\\

\begin{equation}
\begin{aligned}
	\dfrac{\partial U}{\partial r} = -\dfrac{\mu}{r^2} \sum_{n=0}^{\infty} \sum_{m=0}^{n} \left(\dfrac{R_0}{r}\right)^n \left(n+1 \right) \Pbar{n}{m}{u}\\\left(\Cbar{n}{m} \cos\left(m \lambda\right) + \Sbar{n}{m} \sin \left(m \lambda\right) \right)
 \end{aligned}
	\label{eq: dU_dr}
\end{equation}\\

\begin{equation}
\begin{aligned}
	\dfrac{\partial U}{\partial \lambda} = \dfrac{\mu}{r} \sum_{n=0}^{\infty} \sum_{m=0}^{n} \left(\dfrac{R_0}{r}\right)^n \left(n+1 \right) \Pbar{n}{m}{u} m \\\left(-\Cbar{n}{m} \sin\left(m \lambda\right)+\Sbar{n}{m} \cos \left(m \lambda\right) \right)
 \end{aligned}
	\label{eq: dU_dlambda}
\end{equation}

The derivative $\dfrac{\partial U}{\partial \phi}$ is more challenging since it involves the derivatives of the associated Legendre polynomial

\begin{equation}
\begin{aligned}
	\dfrac{\partial U}{\partial \phi} = \dfrac{\mu}{r} \sum_{n=0}^{\infty} \sum_{m=0}^{n} \left(\dfrac{R_0}{r}\right)^n \dfrac{\partial \Pbar{n}{m}{\sin\left(\phi\right)} }{\partial \phi} \\\left(\Cbar{n}{m} \cos\left(m \lambda\right)+ \Sbar{n}{m} \sin \left(m \lambda\right) \right)
 \end{aligned}
	\label{eq: dU_dphi}
\end{equation}

There are several recursive formulas to compute the derivative of the fully normalized Legendre \mbox{polynomials \cite{lundberg1988recursion}}. For this implementation this is achieved by combining the definition of the normalized Legendre polynomials, in Equation \ref{eq: fully-normalized associated legendre polynomial}, with the set of recursive formula in Equation \ref{eq: recursive formula for the computation of P_nm} leading to

\begin{equation}
	\begin{aligned}
		\displaystyle& \dfrac{\partial \Pbar{n}{m}{\sin(\phi)}}{\partial \phi} = \left[\dfrac{\partial \Pbar{n}{m}{u}}{\partial u}\right]  \cos{\phi} = \\
  & \\
  \displaystyle& =  -m \tan{\phi} \Pbar{n}{m}{u} + K\p{n,m} \Pbar{n}{m+1}{u}
  \end{aligned}
	\label{eq: Derivation of dP_nm_dphi 1}
\end{equation}

with 

\begin{equation}
	K_{n,m}  = \left\{ 
	\begin{array}{ c l }
		\sqrt{\p{n-m} \p{n+m+1}} &  m > 0 \\
    & \\
		\sqrt{\dfrac{n \p{n+1}}{2}} &m = 0
	\end{array}
	\right.
	\label{eq: Derivation of dP_nm_dphi 3}
\end{equation}


\section{Normalized coefficients computation}\label{sec: Normalized coefficients computation}
The coefficients $\Cbar{n}{m}$ and $\Sbar{n}{m}$ are a function of the asteroid shape and density distribution. In particular they are obtained integrating a set of trinomial shape functions $\Bar{c}_{nm}$ and $\Bar{s}_{nm}$ over the body \cite{WERNER19971071}

\begin{equation}
\begin{aligned}    
    \begin{bmatrix}
			\Cbar{n}{m} \\ \Sbar{n}{m}
		\end{bmatrix} = & \iiint_B \begin{bmatrix}
			\Bar{c}_{nm}\left(x,y,z\right) \\ \Bar{s}_{sm}\left(x,y,z\right)  
		\end{bmatrix} dm = \\ = &\iiint_B \rho\begin{bmatrix}
			\Bar{c}_{nm}\left(x,y,z\right) \\ \Bar{s}_{sm}\left(x,y,z\right)  
		\end{bmatrix} dx dy dz
  \label{eq: definition of the normalized coefficients}
  \end{aligned}
\end{equation}

 If the integration is performed over a polyhedron, split in a collection of tethraedra, the full integral is equivalent to the summation of the integrals over each \mbox{tetrahedron \cite{panicucci2020uncertainties}}

\begin{equation}
    \hspace{-3 mm}\begin{bmatrix}
			\Cbar{n}{m} \\ \Sbar{n}{m}
		\end{bmatrix} =   \sum_{s = 1}^{n_s} \left( \iiint_s \rho\begin{bmatrix}
			\Bar{c}_{nm}\left(x,y,z\right) \\ \Bar{s}_{sm} \left(x,y,z\right) 
		\end{bmatrix} dx dy dz\right)
  \label{eq: summation over tethraedra}
\end{equation}

Lien and Kajiya \cite{lien1984symbolic} shows that, under the hypothesis of constant density, each of these integrals, can be solved analytically if a proper change of coordinates is performed.\\

\begin{figure}[h!]
	\centering
	\subfloat[]{{	\includegraphics[width=0.5\linewidth]{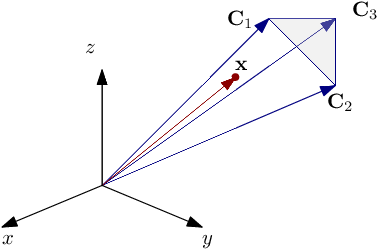}}}%
	\subfloat[]{{	\includegraphics[width=0.5\linewidth]{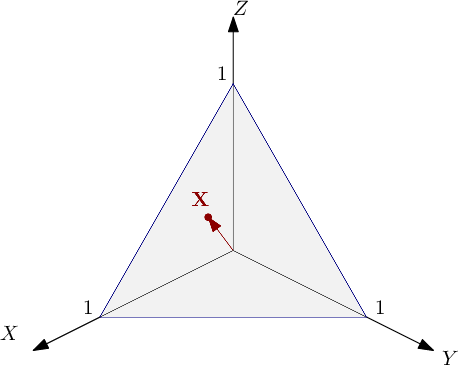}}}%
	\caption{Representation of the single tetrehedron of the polyhedral shape model. (a) Represents the tetrahedron in the asteroid fixed frame $\mathcal{B}$, while (b) shows the standard simplex obtained after the coordinates change.}
	\label{fig: Change of variables}%
\end{figure}

A similar approach is followed here to derive the formulation with variable density distribution. Consider the single polyhedron shown in Figure \ref{fig: Change of variables}a

The general position inside the tetrahedron can be identified as a linear combination of the vertex coordinates $\vec{C}_1$, $\vec{C}_2$ and $\vec{C}_3$ 

\begin{equation}
    \begin{pmatrix}
        x \\ y \\ z  
    \end{pmatrix} =  \begin{bmatrix}
        \vec{C}_1 & \vec{C}_2 & \vec{C}_3
    \end{bmatrix}  \begin{pmatrix}
        X \\ Y \\ Z  
    \end{pmatrix}
    \label{eq: gravity change of variables}
\end{equation}

or better expressed in compact form as

\begin{equation}
     \vec{x} =\vec{J}_s \vec{X}
\end{equation}
where the matrix $\vec{J}_s$ is the Jacobian of the transformation. This change of coordinates allows to perform the integration in Equation \ref{eq: definition of the normalized coefficients} over a standard simplex, as shown in Figure \ref{fig: Change of variables}b. From Equation \ref{eq: summation over tethraedra} thus follows

\begin{equation}
\hspace{-4.1 mm}	\begin{aligned}
		\begin{bmatrix}
			\Cbar{n}{m} \\ \Sbar{n}{m}
		\end{bmatrix} = &\sum_{s = 1}^{n_s} \iiint_{ss} \left( \rho\begin{bmatrix}
			\Bar{c}_{nm}\left(\vec{X}\right) \\ \Bar{s}_{sm} \left(\vec{X}\right) 
		\end{bmatrix} det(\vec{J}_s)\right)  d\vec{X}
	\end{aligned}
 \label{eq: after change of variables}
\end{equation}

Since $\bar{c}_{nm}\left(\vec{x}\right)$ and $\bar{s}_{nm}\left(\vec{x}\right)$ are trinomials of order n, and the transformation in Equation \ref{eq: gravity change of variables} is linear, $\bar{c}_{nm}\left(\vec{X}\right)$ and $\bar{s}_{nm}\left(\vec{X}\right)$ will also be trinomials of order n. The shape functions can then be expressed as 

\begin{equation}
\begin{aligned}    
    \begin{bmatrix}
			\Bar{c}_{nm}\left(\vec{X}\right) \\ \Bar{s}_{sm} \left(\vec{X}\right) 
		\end{bmatrix} = & \sum_{i+j+k = n} \left(\begin{bmatrix}
			\Bar{\alpha}_{ijk} \\ \Bar{\beta}_{ijk} 
		\end{bmatrix} X^i Y^j Z^k\right) = \\
  = & \sum_{i+j+k = n} \left(\vec{p}_{ijk} X^i Y^j Z^k\right)
  \label{eq: trinomials}
  \end{aligned}
\end{equation}

The components $\Bar{\alpha}_{ijk}$ and $\Bar{\beta}_{ijk}$ of vector $\vec{p}_{ijk}$ can be retrieved through a series of recursive relations anchored on certain initial conditions. In this paper the details of this procedure are omitted and the coefficients can be considered to be known. The interested reader can find a comprehensive discussion of this process in \cite{WERNER19971071,panicucci2020uncertainties}. \\

Combining Equations \ref{eq: after change of variables} and \ref{eq: trinomials} the steps in \mbox{Equation \ref{eq: step of the derivatoin before discretizaion}} hold. Note that this does not contain any assumption on the internal density distribution of the body.  

\begin{figure*}[ht!]
\begin{equation}
	\begin{aligned}
		\begin{bmatrix}
			\Cbar{n}{m} \\ \Sbar{n}{m}
		\end{bmatrix} = &
		\sum_{s = 1}^{n_s} \left( det(\vec{J}_s) \iiint_{ss} \rho \sum_{i+j+k = n} \left(\vec{p}_{ijk} X^i Y^j Z^k \right) d\vec{X}\right) = \\ 
		= & \sum_{s = 1}^{n_s} \left( det(\vec{J}_s) \sum_{i+j+k = n}\left(\vec{p}_{ijk} \iiint_{ss} \rho X^i Y^j Z^k d\vec{X}\right)\right) = \\ 
		= & \sum_{s = 1}^{n_s} \left( det(\vec{J}_s) \sum_{i+j+k = n}\left(\vec{p}_{ijk} \int_{0}^{1} \int_{0}^{1-Z} \int_{0}^{1-Z-Y} \rho X^i Y^j Z^k d\vec{X}\right)\right) 
	\end{aligned}
 \label{eq: step of the derivatoin before discretizaion}
\end{equation}
\end{figure*}

\begin{figure}[h!]
	\centering
\includegraphics[width=1\linewidth]{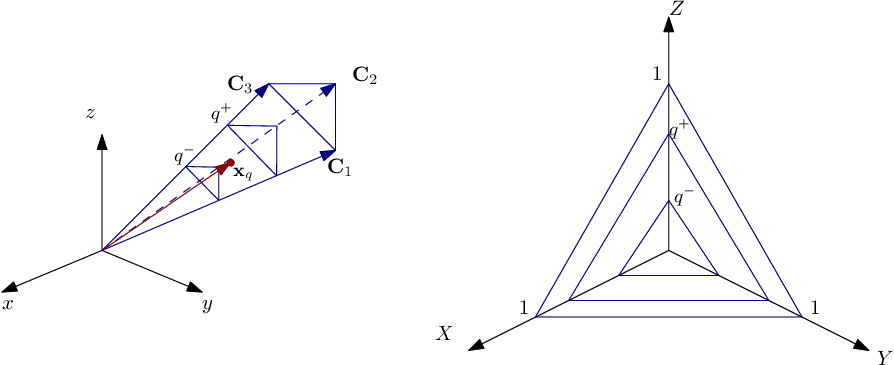}%
	\caption{Radial discretization of the tetrahedron}
	\label{fig: Gravity3}%
\end{figure}

Consider now the uniform discretization of the polyhedron, shown in Figure \ref{fig: Gravity3}, in $n_q$ segments each of them at a constant density $\rho_q = \rho\left(\vec{x}_q\right)$ with $\vec{x}_q$ being the geometric center of segment $q$. 

The integration over the standard simplex can then be split further more in a summation of integrals over each segment, leading to

\begin{equation}
\hspace{-3 mm}\begin{aligned}
  & \begin{bmatrix}
			\Cbar{n}{m} \\ \Sbar{n}{m}
  \end{bmatrix} =   \sum_{s = 1}^{n_s} \left( det(\vec{J}_s) \sum_{i+j+k = n}\left(\begin{bmatrix}
			\Bar{\alpha}_{ijk} \\ \Bar{\beta}_{ijk} 
		\end{bmatrix} \sum_{q=1}^{n_q} \rho_q \right. \right.\\ & \left. \left. \left(\int_{0}^{q^+}   g_{ijk}(Z)  dZ-\int_{0}^{q^-} g_{ijk}(Z) dZ\right)\right)\right)
  \end{aligned}
  \label{eq: second to last step}
\end{equation}

with 

\begin{equation}
    g_{ijk}\left(Z\right) = \int_{0}^{1-Z} \int_{0}^{1-Z-Y}  X^i Y^j Z^k dX dY dZ
    \label{eq: known integral}
\end{equation}

Lien and Kajiya \cite{lien1984symbolic} shows that this integral can be computed analytically as 

\begin{equation}
	g_{ijk}(Z) = \dfrac{\beta\left(j+1,i+2\right)}{i+1} \left(1-Z\right)^{\left(i+j+2\right)}Z^k
\end{equation}

where $\beta$ is the Beta function, also known as Euler's integral, with tabulated values. By substituting the definition of $g_{ijk}(Z)$ in Equation \ref{eq: second to last step} follows that 

\begin{equation}
\begin{aligned}
  \begin{bmatrix}
			\Cbar{n}{m} \\ \Sbar{n}{m}
  \end{bmatrix} = &\sum_{s = 1}^{n_s} \left( det(\Vec{J}_s) \sum_{i+j+k = n}\left(\begin{bmatrix}
			\Bar{\alpha}_{ijk} \\ \Bar{\beta}_{ijk} 
		\end{bmatrix} \right. \right. \\ & \left. \left. \sum_{q=1}^{n_q} \Big(\rho_q \Big(h_{q^+,i,j,k}  \right. \right.
        - h_{q^-,i,j,k}\Big) \Big) \Bigg) \Bigg)
  \end{aligned}
  \label{eq: last step}
\end{equation}

with 

\begin{equation}
	h_{q^+,i,j,k} = \dfrac{\beta\left(j+1,i+2\right)}{i+1} \Tilde{\beta}\left(q,k+1,i+j+3\right)
\end{equation}

and $\tilde{\beta}$ is the incomplete beta function, also known and tabulated.

\section{Validation}\label{sec: Validation}
 The expression provided in Equation \ref{eq: last step} provides an exact way of computing the normalized coefficients $\Cbar{n}{m}$ and $\Sbar{n}{m}$ starting from a generic shape model of the asteroid and a generic density distribution $\rho\left(\vec{x}\right)$. Apart from defining the shape of the gravity field, the normalized coefficients provide also important information on the inertia properties of the body. For example, the Center of Mass (CoM) of the body can be derived as: 

\begin{equation}
    \vec{r}_{CoM,\mathcal{B}} = \begin{bmatrix}
        \Cbar{1}{1} \\ \Sbar{1}{1} \\ \Cbar{1}{0}
    \end{bmatrix} \sqrt{3} R_0
    \label{eq: CoM}
\end{equation}

This can be used to validate the proposed methodology. Five test cases are investigated: the first three are toy problems designed to validate the approach comparing the CoM position computed from the coefficients with known analytical solutions while the other two presents a more realistic application case to asteroid (433) Eros in which the acceleration computed with spherical harmonics model is compared against the one obtained with a Mascon model.\\

\begin{figure} [h!]
	\centering
	\subfloat[]{{	\includegraphics[width=0.8\linewidth]{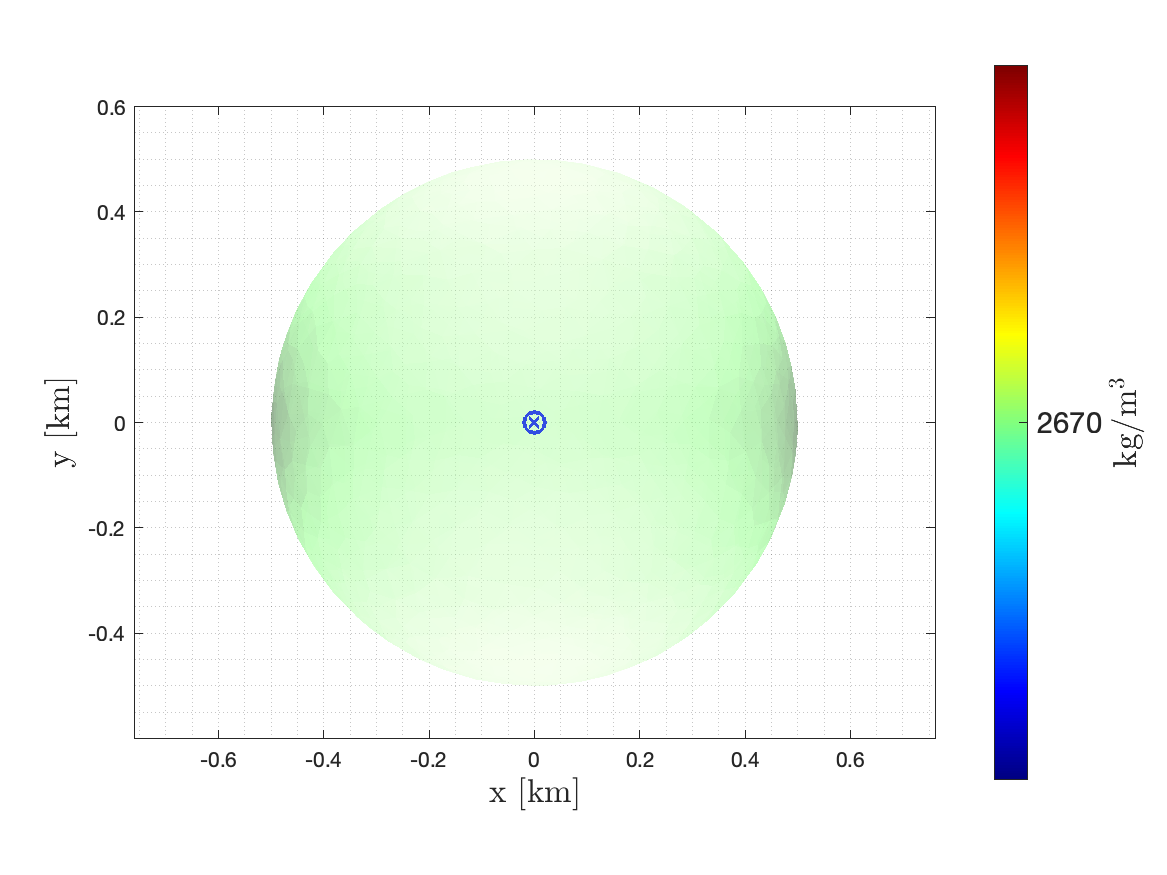}}}%
	\qquad
	\subfloat[]{{	\includegraphics[width=0.8\linewidth]{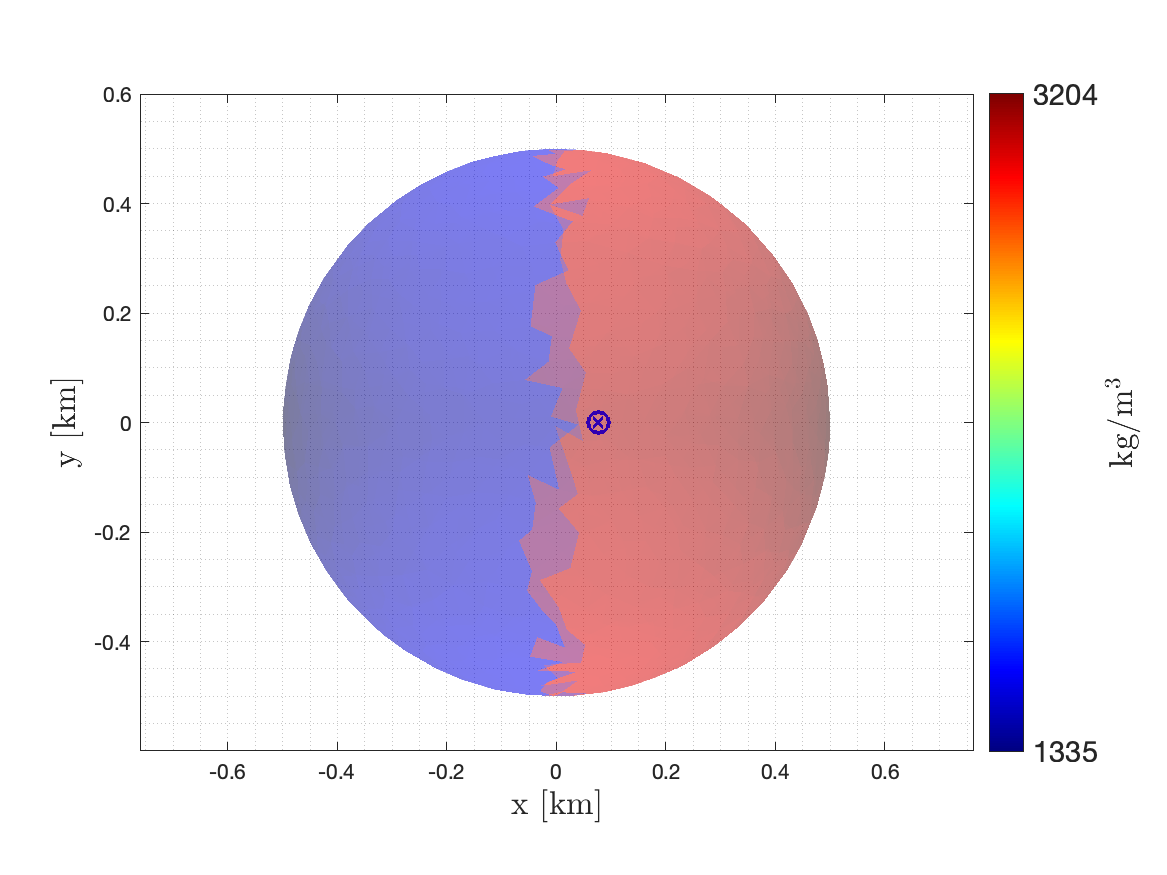}}}%
	\qquad
	\subfloat[]{{	\includegraphics[width=0.8\linewidth]{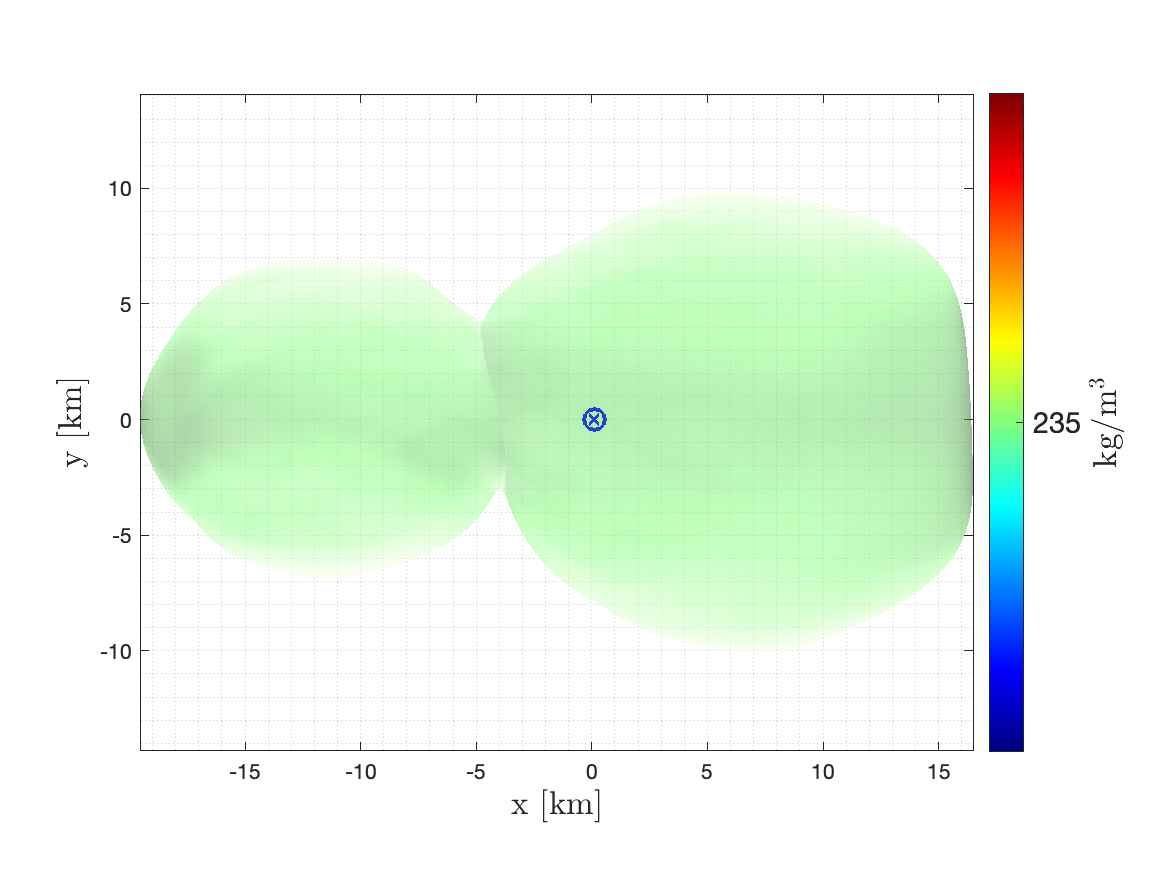}}}%
	\caption{CoM estimation comparison, the cross indicates the one computed with Eq.\ref{eq: CoM} while the circle the reference solution. (a) Uniform density sphere, (b) Variable density sphere, (c) Uniform density model of (486958) Arrokoth asteroid. }%
	\label{fig: CoM location comparison}%
\end{figure}

The first case is the one of a uniform density sphere with density \mbox{2670 kg / m$^3$} and radius $R_0 = 500$ m. Being a sphere with constant density the CoM is clearly in its center, as illustrated in Figure \ref{fig: CoM location comparison}(a). The second case considers the same body but introduces a density gradient between two hemispheres, see Figure \ref{fig: CoM location comparison}(b). In particular the density is assumed to be

\begin{equation}
    \rho \left( \vec{x}\right) = \begin{cases}
    \rho_1 = 3204 \textrm{ kg/m$^3$}& \quad x \ge 0 \\
        \rho_2 = 1335 \textrm{ kg/m$^3$}& \quad x < 0
    \end{cases}
\end{equation}
In this case the CoM will only have an $x$ component that can be analytically computed as 

\begin{equation}
    x_{CoM} = \dfrac{3}{8} R_0 \left( \dfrac{\rho_1 - \rho_2}{\rho_1 + \rho_2}\right)
\end{equation}

The third case considers instead the asteroid (486958) Arrokoth, selected for its peculiar shape, see \mbox{Figure \ref{fig: CoM location comparison}(c)}. A uniform density distribution is assumed here, with a density of $235 \textrm{ kg/m$^3$}$ and a center of mass position \mbox{$\vec{r}_{CoM} = \left[0.101, 0.020, 0.079\right] \textrm{ km} $}, both estimated by Keane et al.\cite{keane2022geophysical}. \\

Normalized spherical harmonics coefficients are computed for these three cases using $n_q = 10$ in the methodology presented in this section, and the CoM is computed as in Equation \ref{eq: CoM}. Table \ref{tab:CoM error} shows the accuracy achieved in the three cases. The table shows that an accuracy always below $1$ meter is achieved even when the CoM location is not analytically computed but retrieved from literature as in the Arrokoth case. 
\begin{table}[h!]
    \centering
    \caption{Accuracy of the proposed model in determining the CoM location.}
    \begin{tabular}{c|c}
    \hline
    \hline
       Case  &  CoM error [m]\\
       \hline
       Uniform sphere  & 0.0162\\
       \hline
       Non-uniform sphere & 0.4217 \\
       \hline
       Arrokoth with uniform density & 0.67 \\
       \hline
       \hline
    \end{tabular}
    \label{tab:CoM error}
\end{table}

The last two examples refers to asteroid Eros. Thanks to extensive mapping performed by NEAR \cite{NEARMissionDesign}, Eros is the small body for which higher fidelity models are available. A good description of its gravity field is provided by Garmier et al. \cite{eros_gravity}. While the study reconstruct the asteroid density to be almost uniform inside the body, gravitational anomalies are detected that could be due to the existence of denser areas in the central \mbox{region \cite{eros_gravity}}.\\

To test the accuracy of the gravity model the acceleration obtained from Equation \ref{eq: central gravity acceleration} is compared with the one computed with a variable density Mascon model \cite{VariableDensityMascon} with the same density distribution assumption. To make the computation affordable, the model is derived starting from a 50k polyhedron model\footnote{\url{https://3d-asteroids.space/asteroids/433-Eros}} and performing a mesh simplification to obtain a polyhedron with 12290 vertices and 24576 faces used to compute the gravitational coefficients. \\

\begin{figure}[h]
    \centering
    \includegraphics[width=0.8\linewidth]{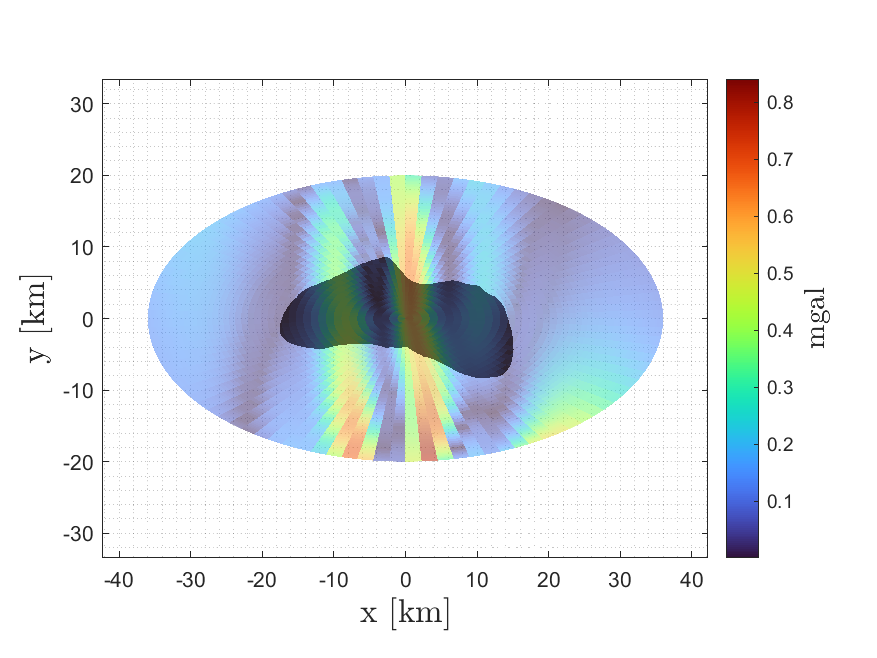}
    \caption{Error in the acceleration obtained comparing the variable density spherical harmonics model discussed in this section with a Mascon model applied to a constant density model of Eros. }
    \label{fig: GravityError Eros Uniform}
\end{figure}

First, a uniform density model is considered with a density 2670 kg/m$^3$. Figure \ref{fig: GravityError Eros Uniform} shows the error in the acceleration among the two models over an ellipsoid with semi-major axis $30\textrm{ km}\times 20\textrm{ km} \times 20\textrm{ km}$ seen from the asteroid north pole. The figure shows an error that is always below $1 \textrm{ mgal} = 1\mathrm{e}{-5}\textrm{ m/s}^2$.  This results to be less than one order of magnitude lower than the perturbation due non spherical gravity at that distance from Eros as shown in Rizza et al. \cite{rizza2024goal}. Also note that the picks in the error are clearly linked with the closest regions to the Brillouin sphere.

Then a variable density distribution with an inner core with a density $10\%$ larger than the nominal values is considered. In particular

\begin{equation}
    \rho \left( \vec{r}\right) = \begin{cases}
    \rho_1 = 2937 \textrm{ kg/m$^3$}& \quad r \le 5 \textrm{ km} \\
        \rho_2 = 2670 \textrm{ kg/m$^3$}& \quad r > 5 \textrm{ km}
    \end{cases}
    \label{eq: variable density profile Eros}
\end{equation}

This density profile, shown in Figure \ref{fig: GravityError Eros Uniform}, leads to an acceleration error that is slightly larger than the one of the previous case, see Figure \ref{fig: GravityError Eros Uniform} but still always below $1 \textrm{ mgal}$.

\begin{figure}[h!]
    \centering
    \includegraphics[width=0.75\linewidth]{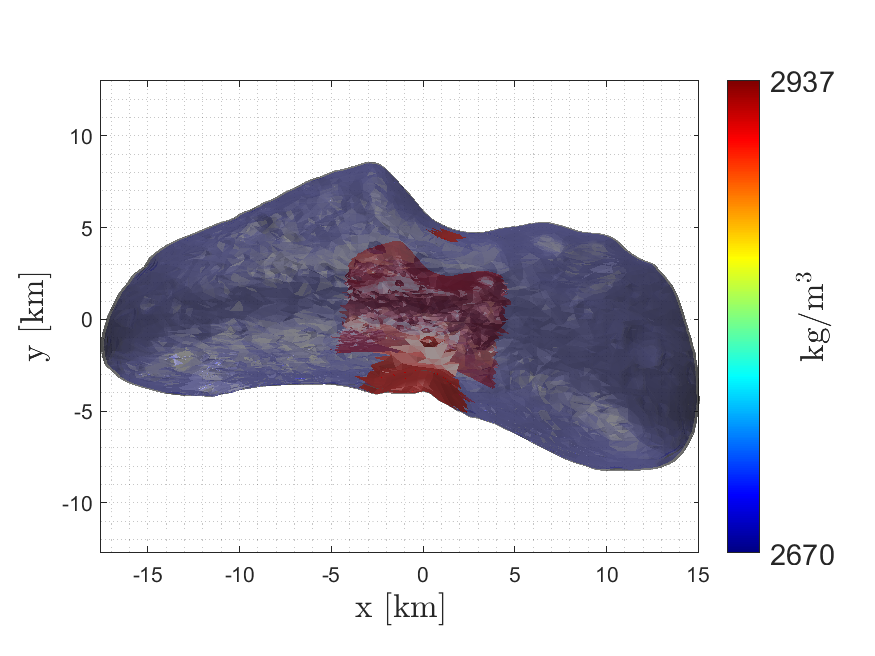}
    \caption{Assumed density distribution for Eros with an heavier inner core.}
    \label{fig: Eros density distribution}
\end{figure}

\begin{figure} [h!]
    \centering
    \includegraphics[width=0.8\linewidth]{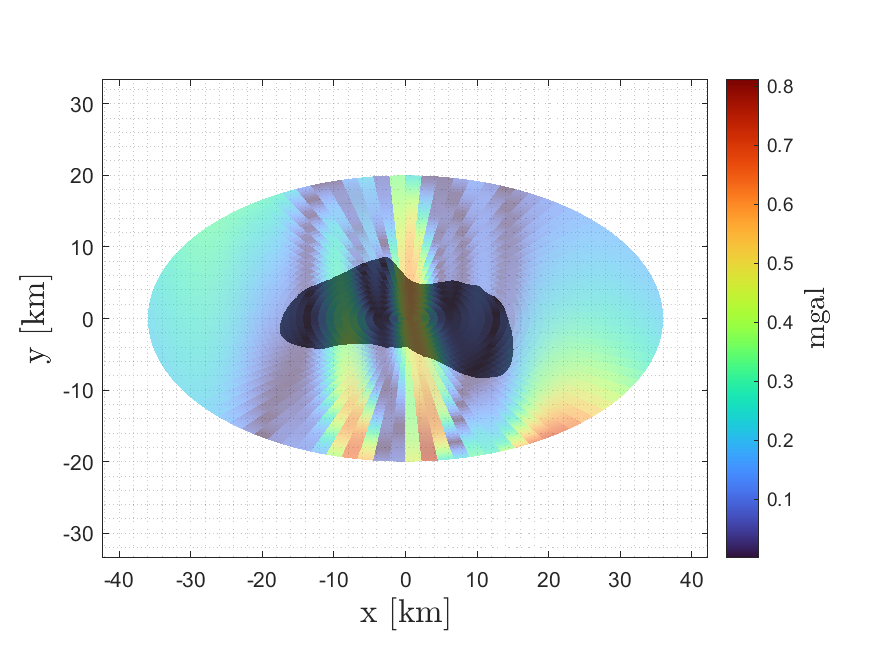}
    \caption{Error in the acceleration obtained comparing the variable density spherical harmonics model discussed in this section with a Mascon model with the internal density distribution shown in Figure \ref{fig: Eros density distribution}. }
    \label{fig: GravityError Eros Variable}
\end{figure}

To show the relevance of density variation on the spacecraft trajectory a comparison is shown in the following. Considering the initial condition shown on \mbox{Table \ref{tab: Initial condition toy problem gravity}}, the trajectory is propagated for 24 h using the gravity field generated from the two previously described density models. \\

\begin{table}[h!]
	\centering
	\caption{Initial condition used to propagate the two trajectories with different gravity models. The spacecraft state is expressed in the inertial frame $\mathcal{N}$.}
	\begin{tabular}{c | c} 
		\hline
		\hline
		$\vec{r}_0$ & $\left[0.71, -45, 0\right]$ km  \\ 
		\hline
		$\vec{v}_0$ & $\left[2.24, -0.035, 2.21\right]$ m/s\\
		\hline
		\hline
	\end{tabular}
	\label{tab: Initial condition toy problem gravity}
\end{table}

\begin{figure} [h!]
	\centering
	\subfloat[]{{	\includegraphics[width=0.8\linewidth]{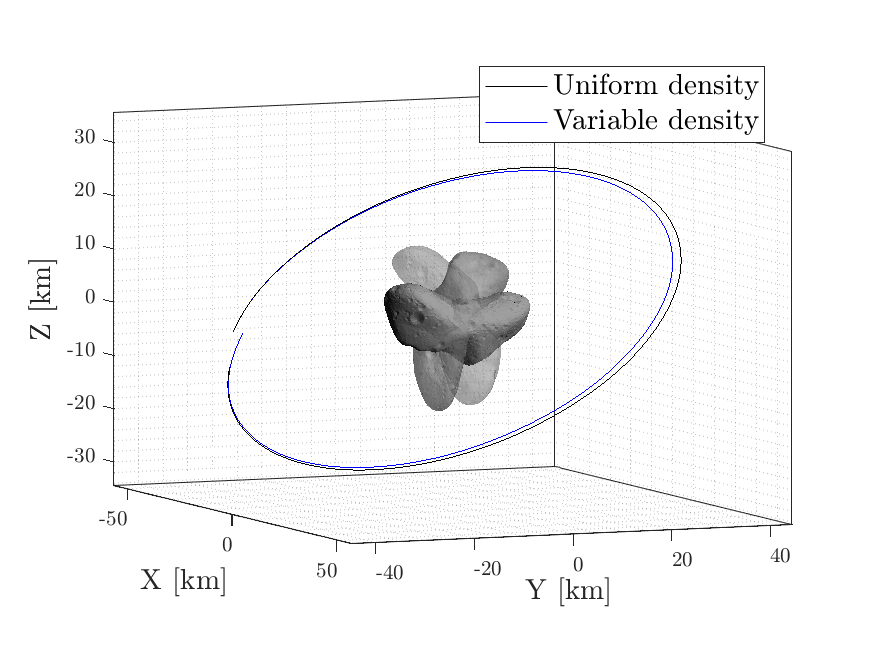}}}%
	\qquad
	\subfloat[]{{	\includegraphics[width=0.8\linewidth]{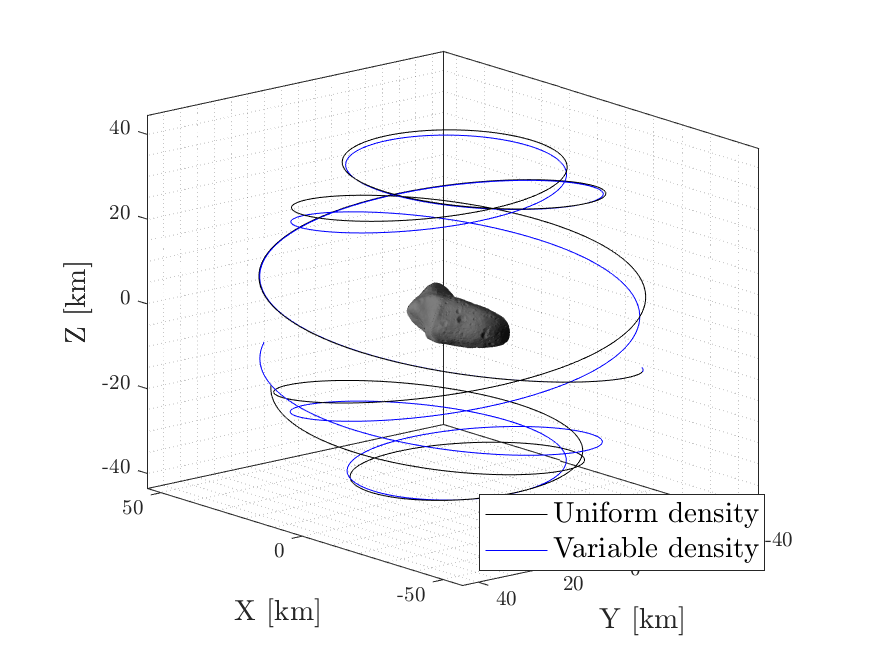}}}%
	\caption{Trajectory comparison with variable density gravity field. (a) Trajectory in the inertial frame $\mathcal{N}$ and, (b) trajectory in the asteroid fixed frame $\mathcal{B}$.}%
	\label{fig:Variable Density - Trajectory comparison}%
\end{figure}

\begin{figure} [h!]
	\centering
	\subfloat[]{{	\includegraphics[width=0.75\linewidth]{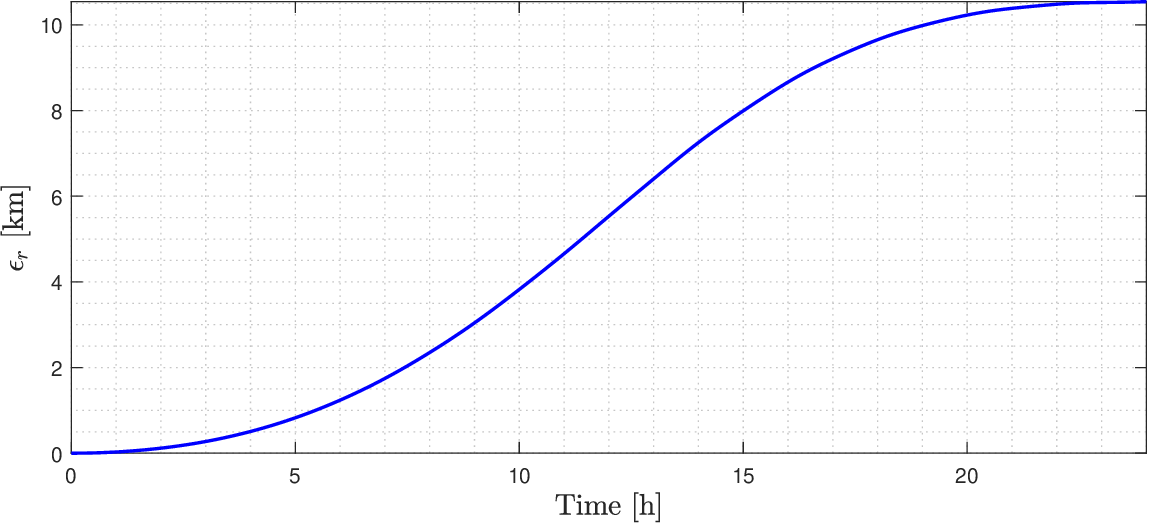}}}%
	 \qquad
	\subfloat[]{{	\includegraphics[width=0.75\linewidth]{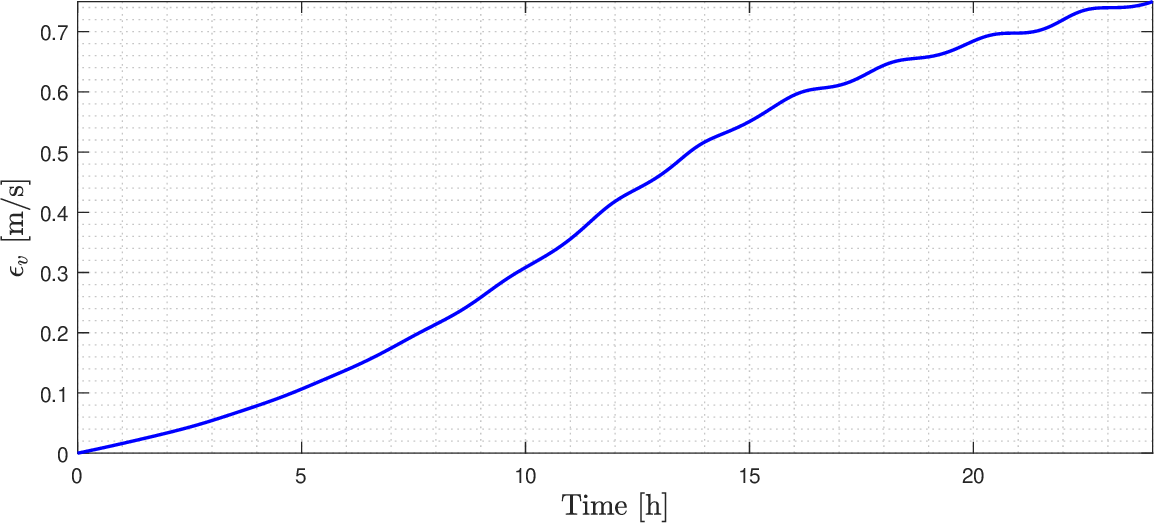}}}%
	\caption{Trajectory error due to variable density gravity field. }%
	\label{fig:Variable Density - Trajectory erros}%
\end{figure}

Inertial and asteroid fixed trajectories are shown in Figure \ref{fig:Variable Density - Trajectory comparison} while the error on position and velocity is reported in Figure \ref{fig:Variable Density - Trajectory erros}.

The observed effect of the variable density gravity field is to induce a cumulative error of more than 10 km and around 0.7 m/s at the end of the considered time horizon. \\

\section{Conclusions}\label{sec: Conclusions}
In this paper a pipeline for semi-analytical computation of normalized spherical-harmonics coefficients is proposed. The algorithm takes as input only a shape model of the asteroid and a generic density distribution function $\rho(\vec{x})$. Different levels of accuracy can be achieved simply by increasing the fidelity of the shape model or the number of radial discretization intervals $n_q$. The methodology has been tested against analytical results, literature findings and numerical simulations, showing good accuracy. The set of generated coefficients could be used on-board to efficiently compute the effects of non-uniform density distribution without relying on computationally heavier gravity models which are difficult to run on-board.  
\section{References}
\printbibliography[heading=none]

@article{arkani1998lunar,
  title={The lunar mascons revisited},
  author={Arkani-Hamed, Jafar},
  journal={Journal of Geophysical Research: Planets},
  volume={103},
  number={E2},
  pages={3709--3739},
  year={1998},
  publisher={Wiley Online Library}
}

@book{scheeres2016orbital,
	title={Orbital motion in strongly perturbed environments: applications to asteroid, comet and planetary satellite orbiters},
	author={D.J. Scheeres},
	note = {Chapter 1-2},
	publisher={Springer Berlin, Heidelberg},
    year = {2012},
	url = {10.1007/978-3-642-03256-1}
}

@article{NEARMissionDesign,
	author = {Dunham, David and McAdams, James and Farquhar, Robert},
	year = {2002},
	month = {01},
	pages = {},
	title = {{NEAR mission design}},
	volume = {23},
	journal = {Johns Hopkins APL Technical Digest (Applied Physics Laboratory)}
}

@article{Dawn_RAYMAN2020233,
	title = {{Lessons from the Dawn mission to Ceres and Vesta}},
	journal = {Acta Astronautica},
	volume = {176},
	pages = {233-237},
	year = {2020},
	issn = {0094-5765},
	doi = {10.1016/j.actaastro.2020.06.023},
	author = {Marc D. Rayman}
}

@incollection{hayabusa,
	title={{The Hayabusa mission}},
	author={Yoshikawa, Makoto and Kawaguchi, Junichiro and Fujiwara, Akira and Tsuchiyama, Akira},
	booktitle={Sample Return Missions},
	pages={123--146},
	year={2021},
	publisher={Elsevier}, 
	doi = {10.1016/B978-0-12-818330-4.00006-9},
		address = {Radarweg 29, PO Box 211, 1000 AE Amsterdam, Netherlands}
}

@article{hayabusa2,
	title={{Hayabusa2 mission overview}},
	author={Watanabe, Sei-ichiro and Tsuda, Yuichi and Yoshikawa, Makoto and Tanaka, Satoshi and Saiki, Takanao and Nakazawa, Satoru},
	journal={Space Science Reviews},
	volume={208},
	number={1},
	pages={3--16},
	year={2017},
	publisher={Springer}, 
	doi = {10.1007/s11214-017-0377-1}
}

@article{RosettaOverview,
	title={{The Rosetta mission: flying towards the origin of the solar system}},
	author={Glassmeier, Karl-Heinz and Boehnhardt, Hermann and Koschny, Detlef and K{\"u}hrt, Ekkehard and Richter, Ingo},
	journal={Space Science Reviews},
	volume={128},
	pages={1--21},
	year={2007},
	publisher={Springer},
	DOI = {10.1007/s11214-006-9140-8}
}

@article{dart,
	title={{AIDA DART asteroid deflection test: Planetary defense and science objectives}},
	author={Cheng, Andrew F and Rivkin, Andrew S and Michel, Patrick and Atchison, Justin and Barnouin, Olivier and Benner, Lance and Chabot, Nancy L and Ernst, Carolyn and Fahnestock, Eugene G and Kueppers, Michael and others},
	journal={Planetary and Space Science},
	volume={157},
	pages={104--115},
	year={2018},
	publisher={Elsevier}, 
	doi = {10.1016/j.pss.2018.02.015}
}

@article{cheng2023momentum,
	title={Momentum transfer from the DART mission kinetic impact on asteroid Dimorphos},
	author={Cheng, Andrew F and Agrusa, Harrison F and Barbee, Brent W and Meyer, Alex J and Farnham, Tony L and Raducan, Sabina D and Richardson, Derek C and Dotto, Elisabetta and Zinzi, Angelo and Della Corte, Vincenzo and others},
	journal={Nature},
	volume={616},
	number={7957},
	pages={457--460},
	year={2023},
	publisher={Nature Publishing Group UK London},
	doi = {10.1038/s41586-023-05878-z}
}

@article{walker2018deep,
	title={{Deep-space CubeSats: thinking inside the box}},
	author={Walker, Roger and Binns, David and Bramanti, Cristina and Casasco, Massimo and Concari, Paolo and Izzo, Dario and Feili, Davar and Fernandez, Pablo and Fernandez, Jesus Gil and Hager, Philipp and others},
	journal={Astronomy \& Geophysics},
	volume={59},
	number={5},
	pages={5--24},
	year={2018},
	publisher={Oxford University Press Oxford, UK}
}

@article{CubeSatsAutonomy,
	title={{Autonomous navigation of an asteroid orbiter enhanced by a beacon satellite in a high-altitude orbit}},
	author={Yin, Weidong and Shi, Yu and Shu, Leizheng and Gao, Yang},
	journal={Astrodynamics},
	pages={1--26},
	year={2024},
	publisher={Springer},
	doi = {10.1007/s42064-023-0172-6}
}

@book{vallado2001fundamentals,
	title={{Fundamentals of astrodynamics and applications}},
	author={Vallado, David A},
	volume={12},
	year={2001},
	publisher={Springer Science \& Business Media},
	isbn = {978-1-881883-14-2}
}

@book{rapp1982fortran,
	title={{A Fortran program for the computation of gravimetric quantities from high degree spherical harmonic expansions}},
	author={Rapp, Richard H},
	year={1982},
	publisher={Ohio State University, Department of Geodetic Science and Surveying},
	doi = {10.21236/ADA123406}
}

@article{lundberg1988recursion,
	title={{Recursion formulas of legendre functions for use with nonsingular geopotential models}},
	author={Lundberg, John B and Schutz, Bob E},
	journal={Journal of Guidance, Control, and Dynamics},
	volume={11},
	number={1},
	pages={31--38},
	year={1988},
	doi = {10.2514/3.20266}
}

@article{WERNER19971071,
title = {{Spherical harmonic coefficients for the potential of a constant-density polyhedron}},
journal = {Computers \& Geosciences},
volume = {23},
number = {10},
pages = {1071-1077},
year = {1997},
issn = {0098-3004},
doi = {10.1016/S0098-3004(97)00110-6},
author = {Robert A. Werner}
}

@article{panicucci2020uncertainties,
  title={{Uncertainties in the gravity spherical harmonics coefficients arising from a stochastic polyhedral shape}},
  author={Panicucci, Paolo and Bercovici, Benjamin and Zenou, Emmanuel and McMahon, Jay and Delpech, Michel and Lebreton, J{\'e}r{\'e}my and Kanani, Keyvan},
  journal={Celestial Mechanics and Dynamical Astronomy},
  volume={132},
  pages={1--27},
  year={2020},
  publisher={Springer},
  doi = {10.1007/s10569-020-09962-8}
}

@article{lien1984symbolic,
%  title={{A symbolic method for calculating the integral properties of arbitrary nonconvex polyhedra}},
%  author={Lien, Sheue-ling and Kajiya, James T},
%  journal={IEEE Computer Graphics and Applications},
%  volume={4},
%  number={10},
%  pages={35--42},
%  year={1984},
%  publisher={IEEE}
%}

@article{keane2022geophysical,
  title={{The geophysical environment of (486958) Arrokoth—A small Kuiper belt object explored by new horizons}},
  author={Keane, James T and Porter, Simon B and Beyer, Ross A and Umurhan, Orkan M and McKinnon, William B and Moore, Jeffrey M and Spencer, John R and Stern, S Alan and Bierson, Carver J and Binzel, Richard P and others},
  journal={Journal of Geophysical Research: Planets},
  volume={127},
  number={6},
  pages={e2021JE007068},
  year={2022},
  publisher={Wiley Online Library},
  doi = {10.1029/2021JE007068}
}

@article{eros_gravity,
author = {Garmier, Romain and Barriot, Jean-Pierre and Konopliv, Alexander S. and Yeomans, Donald K.},
title = {{Modeling of the Eros gravity field as an ellipsoidal harmonic expansion from the NEAR Doppler tracking data}},
journal = {Geophysical Research Letters},
volume = {29},
number = {8},
pages = {72-1-72-3},
doi = {10.1029/2001GL013768},
year = {2002}
}

@inproceedings{rizza2024goal,
	author = {Antonio Rizza and Francesco Topputo and Simone D'Amico},
	title = {{Goal-oriented asteroid mapping under uncertainties using Sequential Convex Programming}},
	booktitle = {AIAA SCITECH 2024 Forum},
	doi = {10.2514/6.2024-1990},
	year = {2024}
}

@article{osiris-rex,
	title={{Digital terrain mapping by the OSIRIS-REx mission}},
	author={Barnouin, OS and Daly, MG and Palmer, EE and Johnson, CL and Gaskell, RW and Al Asad, M and Bierhaus, EB and Craft, KL and Ernst, CM and Espiritu, RC and others},
	journal={Planetary and Space Science},
	volume={180},
	year={2020},
	publisher={Elsevier}, 
	doi = {10.1016/j.pss.2019.104764}
}

@article{chen2019spherical,
  title={Spherical harmonic expansions for the gravitational field of a polyhedral body with polynomial density contrast},
  author={Chen, Cheng and Ouyang, Yongzhong and Bian, Shaofeng},
  journal={Surveys in Geophysics},
  volume={40},
  pages={197--246},
  year={2019},
  publisher={Springer},
  doi = {10.1007/s10712-019-09515-1}
}

@Article{VariableDensityMascon,
AUTHOR = {Antoni, M.},
TITLE = {A review of different mascon approaches for regional gravity field modelling since 1968},
JOURNAL = {History of Geo- and Space Sciences},
VOLUME = {13},
YEAR = {2022},
NUMBER = {2},
PAGES = {205--217},
DOI = {10.5194/hgss-13-205-2022}
}

@article{ScheeresHeterogeneousBennu,
author = {D. J. Scheeres  and A. S. French  and P. Tricarico  and S. R. Chesley  and Y. Takahashi  and D. Farnocchia  and J. W. McMahon  and D. N. Brack  and A. B. Davis  and R.-L. Ballouz  and E. R. Jawin  and B. Rozitis  and J. P. Emery  and A. J. Ryan  and R. S. Park  and B. P. Rush  and N. Mastrodemos  and B. M. Kennedy  and J. Bellerose  and D. P. Lubey  and D. Velez  and A. T. Vaughan  and J. M. Leonard  and J. Geeraert  and B. Page  and P. Antreasian  and E. Mazarico  and K. Getzandanner  and D. Rowlands  and M. C. Moreau  and J. Small  and D. E. Highsmith  and S. Goossens  and E. E. Palmer  and J. R. Weirich  and R. W. Gaskell  and O. S. Barnouin  and M. G. Daly  and J. A. Seabrook  and M. M. Al Asad  and L. C. Philpott  and C. L. Johnson  and C. M. Hartzell  and V. E. Hamilton  and P. Michel  and K. J. Walsh  and M. C. Nolan  and D. S. Lauretta },
title = {Heterogeneous mass distribution of the rubble-pile asteroid (101955) Bennu},
journal = {Science Advances},
volume = {6},
number = {41},
pages = {eabc3350},
year = {2020},
doi = {10.1126/sciadv.abc3350}
}

@article{FERRARI2022114914,
title = {Interior of top-shaped asteroids with cohesionless surface},
journal = {Icarus},
volume = {378},
pages = {114914},
year = {2022},
issn = {0019-1035},
doi = {10.1016/j.icarus.2022.114914},
author = {Fabio Ferrari and Paolo Tanga}
}

@article{werner1996exterior,
  title={{Exterior gravitation of a polyhedron derived and compared with harmonic and mascon gravitation representations of asteroid 4769 Castalia}},
  author={Werner, Robert A and Scheeres, Daniel J},
  journal={Celestial Mechanics and Dynamical Astronomy},
  volume={65},
  pages={313--344},
  year={1996},
  publisher={Springer},
  doi = {10.1007/BF00053511}
}

@article{bercovici2020analytical,
  title={Analytical shape uncertainties in the polyhedron gravity model},
  author={Bercovici, Benjamin and Panicucci, Paolo and McMahon, Jay},
  journal={Celestial Mechanics and Dynamical Astronomy},
  volume={132},
  pages={1--32},
  year={2020},
  publisher={Springer},
doi = {10.1007/s10569-020-09967-3}
}

@inproceedings{tardivel2016limits,
  title={The limits of the mascons approximation of the homogeneous polyhedron},
  author={Tardivel, Simon},
  booktitle={AIAA/AAS Astrodynamics Specialist Conference},
  pages={5261},
  year={2016}
}

@article{Takahashi,
author = {Takahashi, Yu and Scheeres, D. J. and Werner, Robert A.},
title = {Surface Gravity Fields for Asteroids and Comets},
journal = {Journal of Guidance, Control, and Dynamics},
volume = {36},
number = {2},
pages = {362-374},
year = {2013},
doi = {10.2514/1.59144}
}

@inproceedings{buonagura2024,
author = {C. Buonagura and C. Giordano and F. Ferrari and F. Topputo},
year = {2024},
title = {{The Orbital Regime Index: a Comprehensive Parameter to Determine Orbital Regions Around Minor Bodies}},
booktitle = {47th Rocky Mountain AAS
GN\&C Conference},
}

\end{document}